\documentclass[runningheads]{llncs}
\usepackage{graphicx}

\usepackage{tikz}
\usepackage{comment}
\usepackage{amsmath,amssymb} 
\usepackage{color}

\usepackage[width=122mm,left=12mm,paperwidth=146mm,height=193mm,top=12mm,paperheight=217mm]{geometry}

\usepackage{subfiles}
\usepackage{hyperref}
\usepackage{seqsplit}
\usepackage{booktabs}
\usepackage{subcaption}
\def\etal{\emph{et al}. }
\def\ie{\emph{i.e.} }
\def\eg{\emph{e.g.,} }
\hypersetup{
    colorlinks=true,
    linkcolor=blue,
    filecolor=magenta,      
    urlcolor=magenta,
    citecolor=blue,
}

\begin{document}
\pagestyle{headings}
\mainmatter
\def\ECCVSubNumber{738}  

\title{VoxelPose: Towards Multi-Camera 3D Human Pose Estimation in Wild Environment}

\titlerunning{VoxelPose}
\author{Anonymous ECCV submission}
\institute{Paper ID \ECCVSubNumber}

\author{Hanyue Tu\inst{1,2}\thanks{This work is done when Hanyue Tu is an intern at Microsoft Research Asia.} \and
Chunyu Wang\inst{1} \and
Wenjun Zeng\inst{1}}
\authorrunning{Tu \etal}
%
\institute{
Microsoft Research Asia\\
 \email{\{chnuwa,wezeng\}@microsoft.com} \and
 University of Science and Technology of China\\ \email{tuhanyue@mail.ustc.edu.cn}
}

\maketitle

\begin{abstract}
We present \emph{VoxelPose} to estimate $3$D poses of multiple people from multiple camera views. In contrast to the previous efforts which require to establish cross-view correspondence based on noisy and incomplete $2$D pose estimates, \emph{VoxelPose} directly operates in the $3$D space therefore avoids making incorrect decisions in each camera view. To achieve this goal, features in all camera views are aggregated in the $3$D  voxel space and fed into \emph{Cuboid Proposal Network} (CPN) to localize all people. Then we propose \emph{Pose Regression Network} (PRN) to estimate a detailed $3$D pose for each proposal. The approach is robust to occlusion which occurs frequently in practice. Without bells and whistles, it outperforms the previous methods on several public datasets.  The code is available at \href{https://github.com/microsoft/voxelpose-pytorch}{https://github.com/microsoft/voxelpose-pytorch}

\end{abstract}

\section{Introduction}
Estimating $3$D human pose from multiple cameras separated by wide baselines \cite{belagiannis20153d,belagiannis20143d,belagiannis2014multiple,dong2019fast,bridgeman2019multi,qiu2019cross,zhang20204d} has been a longstanding problem in computer vision. The goal is to predict $3$D positions of the landmark joints for all people in a scene. The successful resolution of the task can benefit many applications such as intelligent sports \cite{bridgeman2019multi} and retail analysis.

The previous works such as \cite{dong2019fast,bridgeman2019multi} propose to address the problem in three steps. They first estimate $2$D poses in each camera view independently, for example, by Convolutional Neural Networks (CNN) \cite{cao2017realtime,he2017mask}. Then, in the second step, the poses that correspond to the same person in different views are grouped into clusters according to appearance and geometry cues. The final step is to estimate a $3$D pose for each person (\ie each cluster) by standard methods such as triangulation \cite{hartley2003multiple} or pictorial structure models \cite{amin2013multi}. 

\begin{figure*}
	\centering
	\includegraphics[width=4.8in]{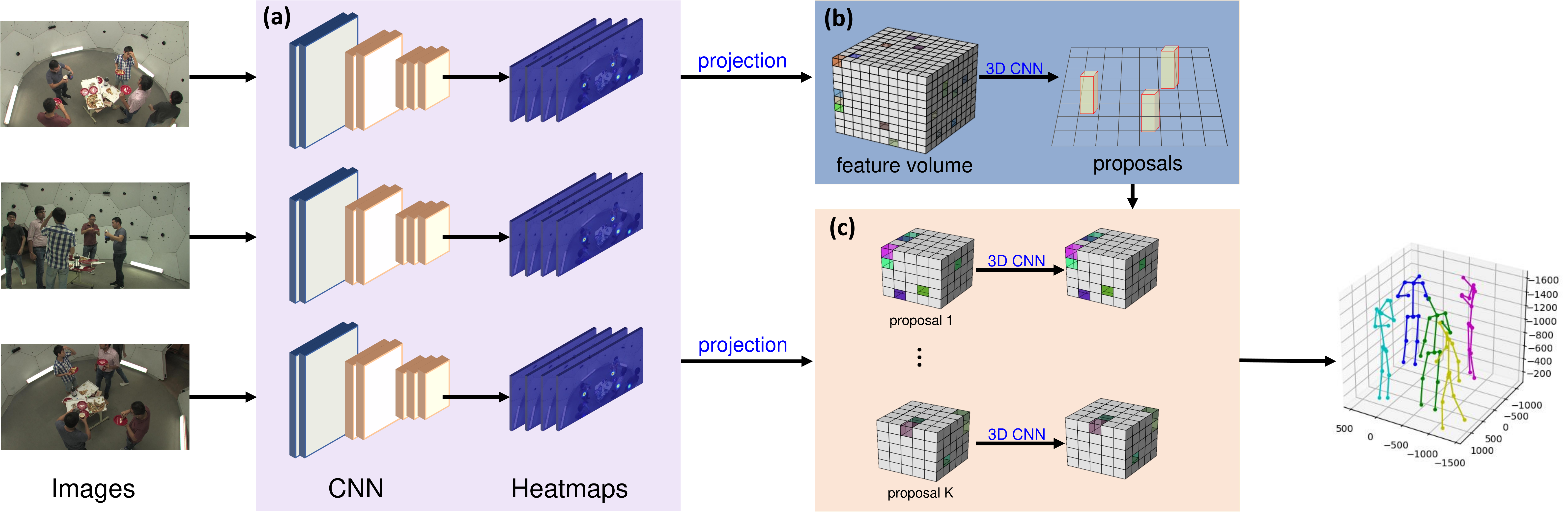}
	\caption{Overview of our approach. It consists of three modules: (a) we first estimate $2$D pose heatmaps for all views; (b) we warp the heatmaps to a common $3$D space and construct a feature volume which is fed into a Cuboid Proposal Network to localize all people instances; (c) for each proposal, we construct a finer-grained feature volume and estimate a $3$D pose.
	}
	\label{fig:pipeline}
\end{figure*}

In spite of the fact that $2$D pose estimation has quickly matured due to the development of CNN models \cite{newell2016stacked,sun2019deep}, the estimation results are still unsatisfactory for challenging cases especially when occlusion occurs which is often the case for natural scenes. See Figure \ref{fig:2dpose} for some example $2$D poses estimated by the state-of-the-art method \cite{sun2019deep}. In addition, it is very difficult to establish cross-view correspondence when $2$D poses are inaccurate. All these pose a serious challenge for $3$D pose estimation in the wild.

To avoid making incorrect decisions for each camera view, we propose a $3$D pose estimator which directly operates in the $3$D space by gathering information from all camera views. Figure \ref{fig:pipeline} shows an overview of our approach. It first estimates $2$D heatmaps for each view to encode per-pixel likelihood of all joints as shown in Figure \ref{fig:pipeline} (a). Different from the previous works, we do not determine the locations of joints (\eg by finding the maximum response) nor group the joints into different instances because estimated heatmaps are usually very noisy and incomplete. Instead, we project the heatmaps of all views to a common $3$D space as in \cite{qiu2019cross} and obtain a more complete feature volume which allows us to accurately estimate the $3$D positions of all joints. 

We first present Cuboid Proposal Network (CPN), as shown in Figure \ref{fig:pipeline} (b), to coarsely localize all people in the scene by predicting a number of $3$D cuboid proposals from the $3$D feature volume. Then for each proposal, we construct a separate \emph{finer-grained} feature volume centered at each proposal, and feed it into a Pose Regression Network (PRN) to estimate a detailed $3$D pose. See Figure \ref{fig:pipeline} (c) for illustration. The two networks are composed of basic $3$D convolution layers and can be jointly trained. 

It is worth noting that our approach implicitly accomplishes two types of association which was previously addressed by post-processing methods. Firstly, the joints of the same person \emph{in a single camera view} are implicitly associated by the cuboid proposal. This was previously addressed in the $2$D space either by bottom-up approaches \cite{cao2017realtime,newell2017associative} or by top-down approaches \cite{sun2019deep} which would suffer when occlusion occurs. Secondly, the joints that correspond to the same person \emph{in different camera views} are also implicitly associated based on the fact that the $2$D poses which overlap with the projections of a $3$D pose belong to the same person. Our approach allows us to avoid the challenging association tasks therefore significantly improves the robustness.

We evaluate our approach on three public datasets including Campus \cite{belagiannis20143d}, Shelf \cite{belagiannis20143d} and CMU Panoptic \cite{Joo_2017_TPAMI}. It outperforms the state-of-the-arts on the first two datasets. Since no work has reported numerical results on Panoptic, we conduct a series of ablation studies by comparing our approach to several baselines. In addition, we find that CPN and PRN can be accurately trained on automatically generated synthetic heatmaps. They achieve similar results as the models trained on realistic images. This is possible mainly because the heatmap based $3$D feature volume representation is a high level abstraction that is disentangled from appearance/lighting, etc. This favorable property dramatically enhances the practical values of the approach.

\section{Related Work}
In this section, we briefly review the related works on $3$D pose estimation for single and multiple people scenarios, respectively. We discuss their main difference from our work and summarize our contributions.

\subsection{Single Person $3$D Pose Estimation}
We briefly classify the existing works into \emph{analytical} and \emph{predictive} approaches based on whether they have learnable parameters. Analytical methods \cite{wang2014robust,qiu2019cross,amin2013multi,ramakrishna2012reconstructing} explicitly model the relationship between a $2$D and $3$D pose according to the camera geometry. \textbf{On one hand}, when multiple cameras are available, the $3$D pose can be fully determined by simple geometry methods such as triangulation \cite{hartley2003multiple} based on the $2$D poses in each view. So the bottleneck lies in the inaccuracy of $2$D pose estimations. Some works \cite{qiu2019cross,amin2013multi} propose to model the conditional dependence between the joints and jointly infer their $3$D positions to improve their robustness to errors in $2$D poses. \textbf{On the other hand}, when only one camera is available, the problem is under-determined because multiple $3$D poses may correspond to the same $2$D pose. The previous works \cite{ramakrishna2012reconstructing,wang2014robust,zhou2016sparse} propose to use low-dimensional pose representations to reduce ambiguities. They optimize the low-dimensional parameters of the  representation such that the discrepancy between its projection and the $2$D pose is minimized. The improvement in $2$D pose estimation has boosted $3$D accuracy.

The predictive models \cite{pavlakos2018ordinal,martinez2017simple,moreno20173d,sun2018integral,fang2018learning,pavllo:videopose3d:2019,bogo2016keep,iskakov2019learnable,remelli2020lightweight} are mainly proposed for the single camera setup aiming to address the ambiguity issue by powerful neural networks. The pioneer works \cite{martinez2017simple,moreno20173d} propose to regress $3$D pose from $2$D joint locations by various networks. Some recent works \cite{qiu2019cross,pavlakos2017coarse,pavlakos2018ordinal,zhou2017towards,iskakov2019learnable} also propose to regress a volumetric $3$D pose representation from images. In particular, in \cite{qiu2019cross,iskakov2019learnable}, the authors project $2$D features or pose heatmaps to $3$D space and estimate $3$D positions of the body joints. The approach achieves better performance than the triangulation and pictorial structure models on \emph{single} person pose estimation. However, it requires to address the challenging association problem in order to apply to scenes with multiple people.

\subsection{Multiple Person $3$D Pose Estimation}
There are two challenging association problems in this task. First, it needs to associate the joints of the same person by either top-down \cite{rogez2019lcr,he2017mask} or bottom-up \cite{cao2017realtime,kreiss2019pifpaf,newell2017associative} strategies. Second, it needs to associate the $2$D poses of the same person in different views based on appearance features \cite{dong2019fast,bridgeman2019multi} which are unstable when people are occluded. The pictorial structure model is extended to deal with multiple people in \cite{belagiannis20143d,belagiannis20153d}. The number of people is assumed to be known which is difficult by itself. Besides, the interactions between different people introduce loops into the graph model and complicate the optimization problem. These challenges limit the $3$D pose estimation accuracy.

Our work differs from the previous methods \cite{dong2019fast,bridgeman2019multi} in that it elegantly avoids the two challenging association problems. This is because a $3$D cuboid proposal already naturally associates the joints of the same person in the same and different views by projecting the proposals to image space. Different from the pictorial structure models \cite{belagiannis20143d,belagiannis20153d}, our approach does not suffer from the local optimum and does not need the number of people in each frame to be known as an input. We find in our experiments that our approach outperforms the previous methods on several public datasets.

\section{Cuboid Proposal Network}
The overview of our approach is shown in Figure \ref{fig:pipeline}. It first estimates $2$D pose heatmaps for every camera view independently by HRNet \cite{sun2019deep}. Then we introduce Cuboid Proposal Network (CPN) to localize all people by a number of cuboid proposals. Finally, we present Pose Regression Network (PRN) to regress a detailed $3$D pose for each proposal. In this section, we focus on the details of CPN including the input, output and network structures.

\subsection{Feature Volume}
\label{sec:featurevolume}
The input to CPN is a $3$D feature volume which contains rich information for detecting people in the $3$D space. The feature volume is constructed by projecting the $2$D pose heatmaps in all camera views to a common discretized $3$D space as will be detailed later. Since the $2$D pose heatmaps encode location information of the joints, the resulting $3$D feature volume also carries rich information for detecting $3$D poses.

\begin{figure*}
	\centering
	\includegraphics[width=4.7in]{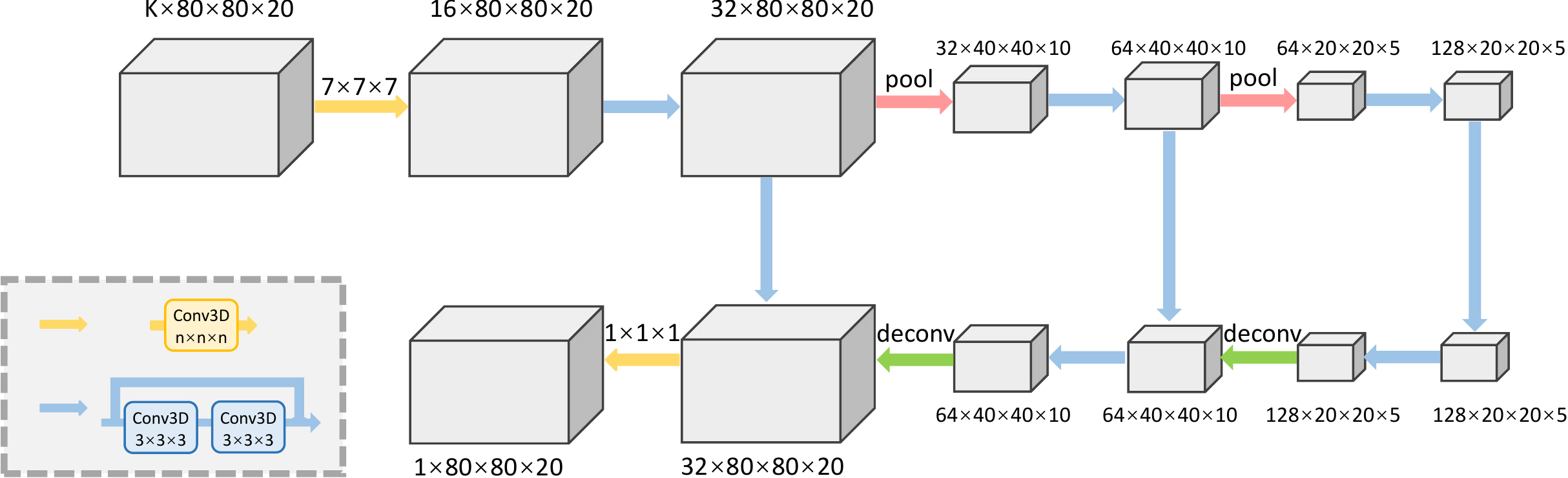}
	\caption{Network structure of CPN. The input is a feature volume (see section \ref{sec:featurevolume}) and the output is the probability map $\mathbf{V}$ (see section \ref{sec:target}). The yellow arrow represents a standard $3$D convolutional layer and the blue arrow represents a Residual Block of two $3$D convolutional layers as shown in the legend.
	}
	\label{fig:cpn}
\end{figure*}

We discretize the $3$D space, in which people can freely move, by $X \times Y \times Z$ discrete locations $\{\mathbf{G}^{x,y,z} \}$. Each location can be regarded as an anchor of people. In order to reduce the quantization error, we set the distance between the neighboring anchors to be small by adjusting the values of $X, Y$ and $Z$, respectively. In general, the space is about $8\text{m} \times 8\text{m} \times 2\text{m}$ on the public datasets \cite{Joo_2017_TPAMI,belagiannis20143d}. So we set $X$, $Y$ and $Z$ to be $80$, $80$ and $20$, respectively, to strike a good balance between speed and precision. The distance between two neighboring bins is about $100$mm which is sufficiently accurate for coarsely localizing people. Note that we will obtain finer-grained $3$D poses in PRN.

We compute a feature vector for each anchor by sampling and fusing the $2$D heatmap values at its projected locations in all camera views. Denote the $2$D heatmap of view $v$ as $\mathbf{M}_v \in \mathcal{R}^{K \times H \times W}$ where $K$ is the number of body joints. For each anchor location $\mathbf{G}^{x,y,z}$, we compute its projected location in view $v$ as $\mathbf{P}^{x,y,z}_{v}$. The heatmap values at $\mathbf{P}^{x,y,z}_{v}$ is denoted as $\mathbf{M}_v^{x,y,z} \in \mathcal{R}^K$. Then we compute a feature vector for the anchor as the average heatmap values in all camera views: $\mathbf{F}^{x,y,z} = \frac{1}{V} \sum_{v=1}^{V}{\mathbf{M}_v^{x,y,z}}$ where $V$ is the number of cameras. More advanced fusion strategies, for example, assigning a data-dependent weight to reflect the heatmap estimation quality in each camera view, could be explored in the future work. In this work, we stick to the the approach of computing average in order to keep the overall approach as simple as possible. We can see that $\mathbf{F}^{x,y,z}$ actually encodes the likelihood that the $K$ joints are at $\mathbf{G}^{x,y,z}$ which is sufficient to infer people presence. In the following sections, we will describe how we estimate cuboid proposals from the feature volume $\mathbf{F}$.

\subsection{Cuboid Proposals}
\label{sec:target}
We represent a cuboid proposal by a $3$D bounding box whose orientation and size are fixed in our experiments. This is a reasonable simplification because the sizes of people in \emph{$3$D space} have limited variations which differs from $2$D proposals in object detection \cite{ren2015faster}. So the main task in CPN is to determine the people presence likelihood at each anchor location.

\begin{figure*}
	\centering
	\includegraphics[width=4.8in]{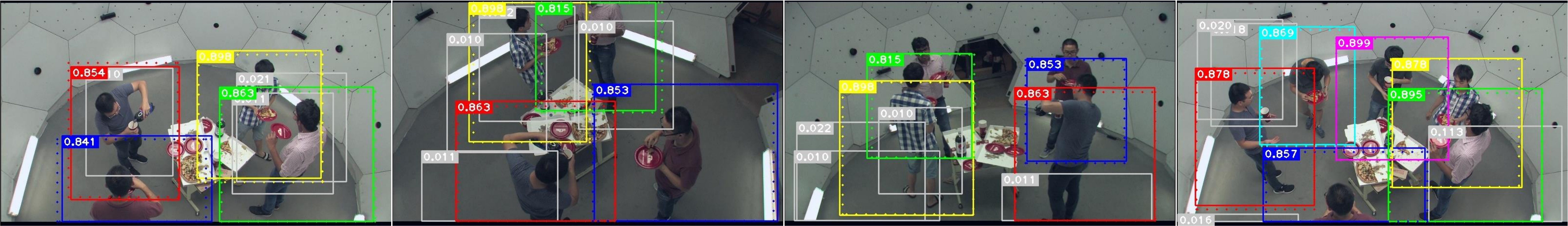}
	\caption{Estimated cuboid proposals. We project the eight corners of each proposal to the image and compute the minimum and maximum coordinates along the x and y-axis, respectively, which form a bounding box. The numbers represent the estimated confidence scores. The gray boxes denote low confidence proposals. The dashed boxes are the ground-truth.
	}
	\label{fig:proposal}
\end{figure*}

To generate proposals, we slide a small network over the feature volume $\mathbf{F}$. Each sliding window centered at an anchor is mapped to a lower-dimensional feature which is fed to a fully connected layer to regress a confidence score $\mathbf{V}^{x,y,z}$ representing the likelihood of people presence at the location. The likelihood at all anchors form a $3$D heatmap $\mathbf{V}  \in \mathcal{R}^{X,Y,Z}$. Since we use fixed box orientation and size, we do not estimate them as the $2$D object detectors \cite{ren2015faster,redmon2018yolov3}. Neither do we estimate center offsets relative to the anchor locations for precise people locations because coarse locations are sufficient.

We compute a Ground-Truth (GT) heatmap score $\mathbf{V}_{*}^{x,y,z}$ for every anchor according to its distance to the GT poses. Specifically, for each pair of GT pose (root joint) and anchor, we compute a Gaussian score according to their distance. The score decreases exponentially when the distance increases. Note that there could be multiple scores for one anchor if there are multiple people in the scene and we simply keep the largest one. We train CPN by minimizing:
\begin{equation}
    \mathcal{L_{\text{CPN}}}=\sum_{x=1}^{X} \sum_{y=1}^{Y} \sum_{z=1}^{Z} {\|\mathbf{V}_{*}^{x,y,z} - \mathbf{V}^{x,y,z}\|_2}
\end{equation}
The edge length of every proposal is set to be $2000$mm which is sufficiently large to cover people in arbitrary poses. The orientations of the proposals are aligned with the world coordinate axes.

\subsection{Non-Maximum Suppression}

 We select the anchors with large regression confidence values as the proposals. On top of the 3D heatmap, we perform Non-Maximum Suppression (NMS) based on the heatmap scores to extract local peaks. Then, we keep the locations of peaks whose heatmap scores are larger than a threshold. Similar to \cite{zhou2019objects}, IOU based NMS is not needed for generating proposals because we only have one positive anchor per pose.

\subsection{Network Structures of CPN}
Inspired by the \emph{voxel-to-voxel} prediction network in \cite{moon2018v2v}, we also adopt the $3$D convolutions as the basic building blocks for CPN. Since the input feature volume is sparse and has clear semantic meanings, we propose a simpler structure than \cite{moon2018v2v} which is shown in Figure \ref{fig:cpn}. In some scenarios such as football court, the motion capture space could be larger than that of the public datasets, which would lead to larger feature volume, thus notably reducing the inference speed. We solve the problem by using sparse $3$D convolutions \cite{yan2018second} because the feature volume only has a small number of non-zero values.

We visualize some estimated proposals in Figure \ref{fig:proposal}. We project the $3$D proposals to $2$D images using the camera parameters for the sake of simplicity. We can see that most people instances can be accurately retrieved even though some of them are severely occluded in the current view. This is mainly due to the effective fusion of multiview features in a common $3$D space.  We will numerically evaluate CPN in more details in the experiment section.

\section{Pose Regression Network}
In this section, we present the details of Pose Regression Network (PRN) which, for each proposal, predicts a complete $3$D pose. 

\subsection{Constructing Feature Volume}
Recall that we have already constructed a big feature volume in the previous CPN step, which covers the whole motion space, to coarsely localize people in the environment. However, we do NOT reuse it here in PRN because it is too coarse to accurately estimate the $3$D positions of all joints. Instead, we construct a separate finer-grained feature volume centered at each proposal. The size of the feature volume is set to be $2000\text{mm} \times 2000\text{mm} \times 2000\text{mm}$, which is much smaller than that of CPN ($8\text{m} \times 8\text{m} \times 2\text{m}$), but is still large enough to cover people in arbitrary poses. This volume is divided into a discrete grid with $X' \times Y' \times Z'$ bins where $X'=Y'=Z'=64$. The edge length of a bin is about $\frac{2000}{64}=31.25\text{mm}$. Note that the precision of our approach is not bounded to $31.25$mm because we will use the integration trick \cite{sun2018integral} to reduce the impact of quantization error as will be described in more detail later. With these definitions, we compute the feature volume following the descriptions in section \ref{sec:featurevolume}.

\subsection{Regression of Human Poses}
We estimate a $3$D heatmap $\mathbf{H}_k \in \mathcal{R}^{X' \times Y' \times Z'}$ for each joint $k$ based on the constructed feature volume.  Then the $3$D location $\mathbf{J}_k$ of the joint can be obtained by computing the center of mass of $\mathbf{H}_k$ according to the following formula:
\begin{equation}
\label{formula:integration}
    \mathbf{J}_k = \sum_{x=1}^{X'}\sum_{y=1}^{Y'}\sum_{z=1}^{Z'} (x, y, z) \cdot {\mathbf{H}_k}(x,y,z)
\end{equation}
Note that we do not obtain the location $\mathbf{J}_k$ by finding the maximum of $\mathbf{H}_k$ because the quantization error of $31.25$mm is still large. Computing the expectation as in the above equation effectively reduces the error. This technique is frequently used in the previous works such as \cite{sun2018integral}.

The estimated joint location is compared to the ground-truth location $\mathbf{J}_{*}$ to train PRN. Specifically, the $L_1$ loss is used:
\begin{equation}
    \mathcal{L_{\text{PRN}}}=\sum_{k=1}^{K} {\|\mathbf{J}_{*}^{k} - \mathbf{J}^{k}\|_1}
\end{equation}

The network of PRN is kept the same as CPN as shown in Figure \ref{fig:cpn} except that the input and output are different. The network weights are shared for different joints. We conducted experiments using different weights but that did not make much difference on current datasets. 

\subsection{Training Strategies}
We first train the $2$D pose estimation network for $20$ epochs. The initial learning rate is 1e-4, and decreases to 1e-5 and 1e-6 at the $10_{th}$ and $15_{th}$ epochs, respectively. Then we jointly train the whole network including CPN and PRN for $10$ epochs to convergence. The learning rate is set to be 1e-4. In some experiments (which will be described clearly), we directly use the backbone network learned on the COCO dataset without finetuning on target datasets.

\section{Datasets and Metrics}
\label{sec:datasets}
\textbf{The Campus Dataset \cite{belagiannis20143d}}
This dataset captures three people interacting with each other in an outdoor environment by three cameras. We follow \cite{belagiannis20143d,dong2019fast} to split the dataset into training and testing subsets. To avoid over-fitting to this small training data, we directly use the $2$D pose estimator trained on the COCO dataset and only train CPN and PRN.

\noindent
\textbf{The Shelf Dataset \cite{belagiannis20143d}}
It captures four people disassembling a shelf by five cameras. Similar to what we do for Campus, we use the $2$D pose estimator trained on COCO and only train CPN and PRN.

\noindent
\textbf{The CMU Panoptic Dataset \cite{Joo_2017_TPAMI}
}
It captures people doing daily activities by dozens of cameras among which five HD cameras (3, 6, 12, 13, 23) are used in our experiments. We also report results for fewer cameras. Following \cite{xiang2019monocular}, the training set consists of the following sequences: \seqsplit{``160422\_ultimatum1'', ``160224\_haggling1'', ``160226\_haggling1'', ``161202\_haggling1'', ``160906\_ian1'', ``160906\_ian2'', ``160906\_ian3'', ``160906\_band1'', ``160906\_band2'', ``160906\_band3''.} The testing set consists of :\seqsplit{ ``160906\_pizza1'', ``160422\_haggling1'', ``160906\_ian5'', and ``160906\_band4''}.

\begin{figure}[ht]
\centering
\begin{subfigure}{.29\textwidth}
  \centering
  \includegraphics[width=.99\linewidth]{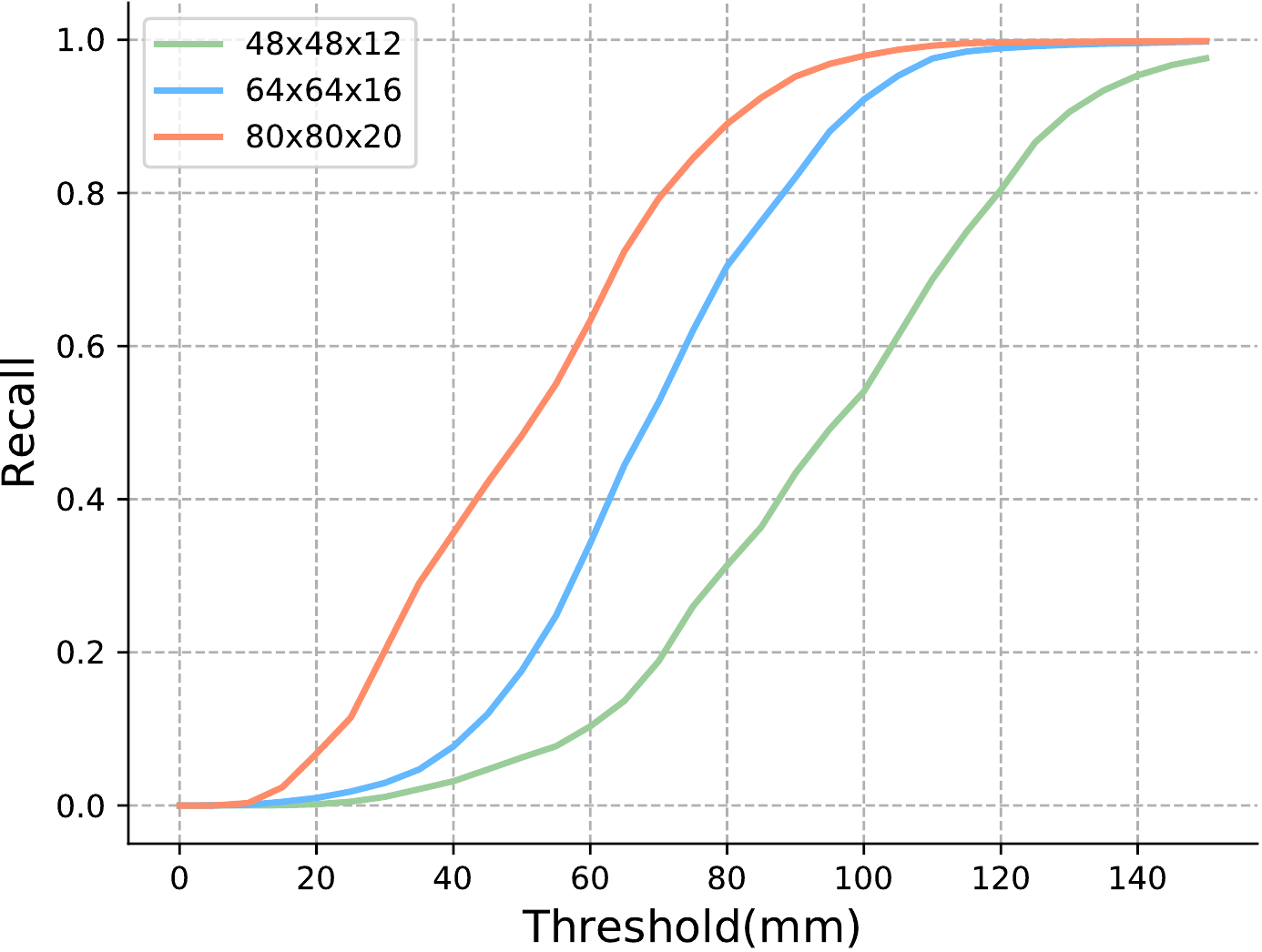}  
  \caption{}
  \label{fig:sub-first}
\end{subfigure}
\begin{subfigure}{.29\textwidth}
  \centering
  \includegraphics[width=.99\linewidth]{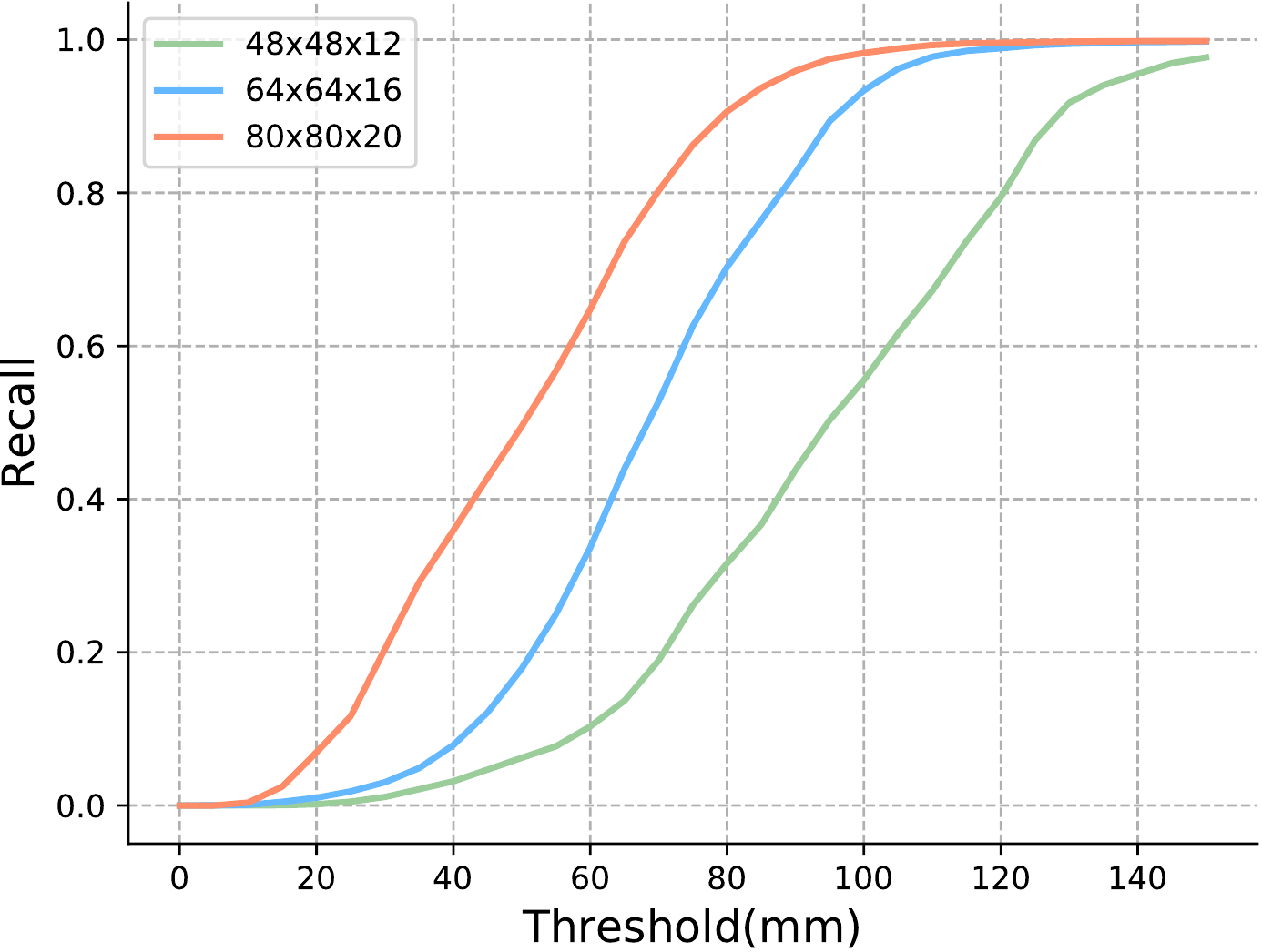}  
  \caption{}
  \label{fig:sub-second}
\end{subfigure}
\begin{subfigure}{.29\textwidth}
  \centering
  \includegraphics[width=.99\linewidth]{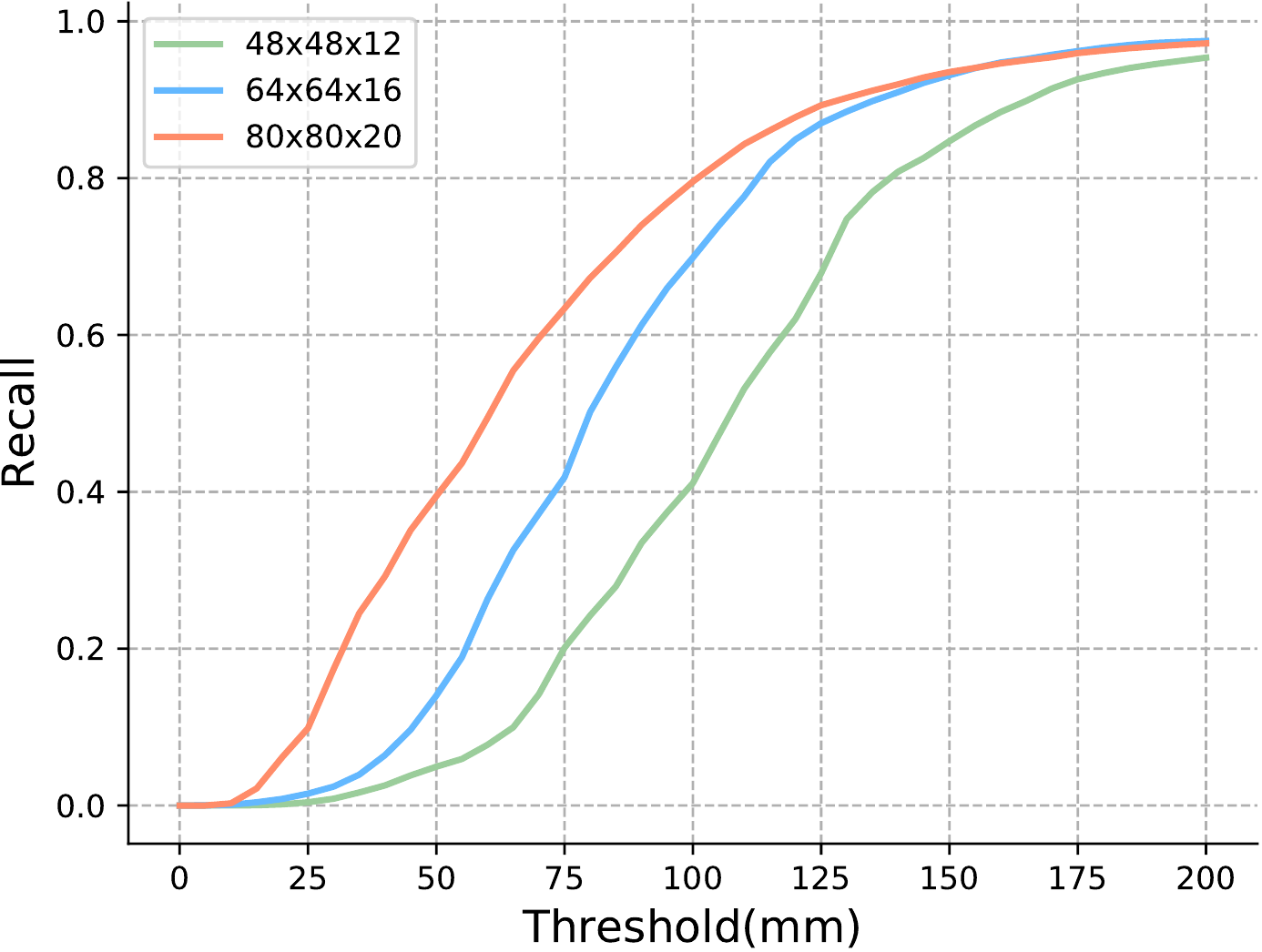}  
  \caption{}
  \label{fig:sub-third}
\end{subfigure}
\caption{Recall curves when the motion space is discretized with different numbers of bins on the Panoptic dataset. \textbf{(a)} CPN is trained/tested on the real images of five cameras from the Panoptic dataset. \textbf{(b)} CPN is trained on the synthetic heatmaps and tested on the real images of five cameras. \textbf{(c)} CPN is trained/tested on the real images of one camera from the Panoptic dataset.}
\label{fig:recall}
\end{figure}

\noindent
\textbf{The Proposal Evaluation Metric}
We compute recall of proposals at different proposal-groundtruth-distance. It is noteworthy that this metric is only loosely related to the $3$D estimation accuracy. We keep ten proposals after NMS for evaluation on the three datasets.

\noindent
\textbf{The $3$D Pose Estimation Metric} Following \cite{dong2019fast}, we use the Percentage of Correct Parts (PCP3D) metric to evaluate the estimated $3$D poses. Specifically, for each ground-truth $3$D pose, it finds the closest pose estimation and computes the percentage of correct parts. This metric does not penalize false positive pose estimations. To overcome the limitation, we also extend the Average Precision (AP$_{K}$) metric \cite{pishchulin2016deepcut} to the multi-person $3$D pose estimation task which is more comprehensive than PCP3D. If the Mean
Per Joint Position Error (MPJPE) of a pose is smaller than $K$ millimeters, we think the pose is accurately estimated.

\section{Evaluation of CPN}
We first study the impact of the space division granularity to the proposals by setting different values to the $X, Y$ and $Z$ parameters. The results are shown in Figure \ref{fig:recall} (a). When we increase the number of bins from $48 \times 48 \times 12$ to $80 \times 80 \times 20$, the recall improves significantly for small thresholds. This is mainly because the quantization error is effectively reduced and the locations of the proposals become more precise.  However, the gap becomes smaller for large thresholds. In our experiments, we use $80 \times 80 \times 20$ bins to strike a good balance between the accuracy and speed.

We also consider a practical situation where we do not have sufficient data to train CPN. We propose to address the problem by generating many synthetic heatmaps: we place a number of $3$D poses (sampled from the motion capture datasets) at random locations in the space and project them to all views to get the respective $2$D locations. Then we generate $2$D heatmaps from the locations to train CPN. The experimental results are shown in Figure \ref{fig:cpn} (b). We can see that the performance is on par with the model trained on the real images. This significantly improves the general applicability of CPN in the wild (we may also need to address the domain adaptation problem in $2$D heatmap estimation but it is beyond the scope of this work).

\begin{table*}[]
\setlength{\tabcolsep}{15pt}
    \centering
    \begin{tabular}{r||ccc}
        \toprule
         Methods &  AP & AP$^{50}$ & AP$^{75}$ \\
         \midrule
         HRNet-w48\cite{sun2019deep} & 55.8 & 67.4 & 59.0 \\
         Ours & \textbf{98.3} & \textbf{99.5} & \textbf{99.1} \\
         \bottomrule
    \end{tabular}
    \caption{2D pose estimation accuracy on the Panoptic dataset. Ours are obtained by projecting the estimated $3$D poses to the images.}
    \label{tab:panoptic_2d}
\end{table*}

Finally, we study the impact of the number of cameras to the proposals. In general, the recall decreases when fewer cameras are used. In particular, the results of a single camera are shown in Figure \ref{fig:recall} (c). We can see that the recall rates at different thresholds are consistently lower than those of the five-camera setup in (a). However, it is still larger than $95\%$ at the threshold of $175$mm. It means that it can coarsely retrieve most people using a single camera which demonstrates its practical feasibility. We will report the ultimate $3$D pose error using a single camera in the next section.

\section{Evaluation of PRN}
\subsection{2D Pose Estimation Accuracy}
\begin{figure*}
	\centering
	\includegraphics[width=4.6in]{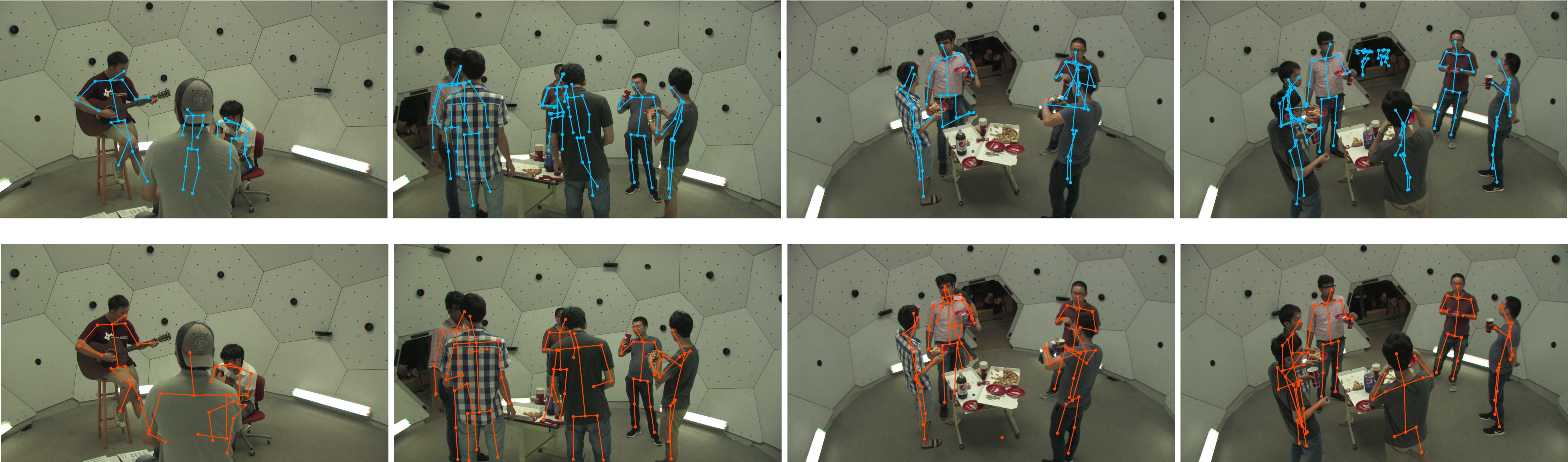}
	\caption{Comparison of $2$D poses estimated by HRNet \cite{sun2019deep} (top row) and our approach (bottom row). Ours are obtained by projecting the $3$D poses to the images. Note that this is only proof-of-concept result rather than rigorous fair comparison as our approach uses multiview images as input. }
	\label{fig:2dpose}
\end{figure*}

We project the $3$D poses estimated by our approach to $2$D and compare them to the results of HRNet \cite{sun2019deep}. Since our approach also uses HRNet to estimate heatmaps, they are comparable. The two models are both trained on the Panoptic dataset \cite{Joo_2017_TPAMI}. The results are shown in Table \ref{tab:panoptic_2d}. The AP of HRNet is only $55.8\%$ because there is severe occlusion. 
Figure \ref{fig:2dpose} shows some $2$D poses estimated by HRNet and our approach, respectively. HRNet gets accurate estimates when there is no occlusion which validates its effectiveness. However, it gets inaccurate estimates when people are occluded. In addition, as a top-down method, it may generate false positives if object detector fails. For example, there are two poses mistakenly detected at the dome entrance area in the fourth example.

\subsection{Ablation Study on $3$D Pose Estimation}
We conduct ablation studies to evaluate a variety of factors of our approach. The results on the Panoptic dataset are shown in Table \ref{tab:ablative_panoptic}. 

\paragraph{\textbf{Space Division Granularity of CPN}}
By comparing the results of (a) and (b), we can see that increasing the number of bins from $64\times64\times16$ to $80\times80\times20$ improves accuracy in general. In particular, the AP$_{25}$ metric improves most significantly whereas AP$_{50}$ improves only marginally. The results represent that using finer-grained grids improves the precision but not accuracy which agrees with our expectation. Further increasing the grid size only slightly decreases the error but notably increases the computation time. To strike a good balance, we use $80\times80\times20$ for the rest of the experiments.

\paragraph{\textbf{Number of Cameras}}
As shown in (b-d) of Table \ref{tab:ablative_panoptic}, reducing the number of cameras generally increases the $3$D error because the information in the feature volume becomes less complete. In extreme cases, when there is only one camera, the $3$D error increases dramatically to $66.95$mm as shown in row (d). This is mainly because there is severe ambiguity in monocular $3$D pose estimation. If we align the pelvis joints of the estimated poses to the ground-truth (as the previous methods), the $3$D pose error decreases to $51.14$mm as shown in Table \ref{tab:ablative_panoptic} (j). This is comparable to the state-of-the-art monocular $3$D pose estimation methods such as \cite{martinez2017simple,ci2019lcn,iskakov2019learnable}. In addition, we find that AP$_{25}$ drops dramatically but AP$_{150}$ only drops slightly when we reduce the number of cameras from five to one. This means that it can estimate coarse $3$D poses using a single camera although they are not as precise as the multiview setup.

\paragraph{\textbf{Generalization to Different Cameras}} We  train and test our approach on different sets of cameras. Specifically, we randomly select a few cameras from the remaining HD cameras for training and test on the selected five cameras. The $3$D error is about $25.51$mm (f) which is larger than the situation where training and testing are on the same cameras. But this is still a reasonably good result demonstrating that the approach has strong generalization capability.

\paragraph{\textbf{Impact of Heatmaps}}
By comparing the results in (b) and (g), we can see that getting accurate $2$D heatmaps is critical to the $3$D accuracy. When the heatmaps are the ground-truth, the $AP$s at a variety of thresholds are very high suggesting that the estimated poses are accurate. The MPJPE remarkably decreases to $11.77$mm. The main reason for this remaining small error is the quantization error caused by space discretization.

\begin{table*}[]
    \setlength{\tabcolsep}{3pt}
    \centering
    \begin{tabular}{ccccccccc}
    \toprule
    & \# Views  & Backbone & CPN Size & AP$_{25}$ & AP$_{50}$ & AP$_{100}$ & AP$_{150}$ & MPJPE \\
    \toprule
    (a) & 5 & ResNet-50 & $64\times 64 \times 16$ & 81.54 & 98.24 & 99.56 & 99.85 & 18.15mm \\
    (b) & 5 & ResNet-50 & $80\times 80 \times 20$ & 83.59 & 98.33 & 99.76 &  99.91 & 17.68mm \\
    (c) & 3 & ResNet-50 & $80\times 80 \times 20$ & 58.94 & 93.88 & 98.45 &  99.32 & 24.29mm \\
    (d) & 1 & ResNet-50 & $80\times 80 \times 20$& 0.860 & 23.47 & 80.69 &  93.32 & 66.95mm \\
    \midrule
    (e)$^{*}$ & 5 & ResNet-50 & $80\times 80 \times 20$ & 71.26 & 96.96 & 99.12 & 99.52 & 20.31mm \\
    (f)$^{+}$ & 5 & ResNet-50 & $80\times 80 \times 20$ & 50.91 & 95.25 & 99.36 & 99.56 & 25.51mm \\
    \midrule
    (g) & 5 & GT Heatmap & $80\times 80 \times 20$ & 98.61 & 99.82 & 99.98 & 99.99 & 11.77mm \\
    (h) & 5 & ResNet-50 & GT Proposal & - & - & - & - & 16.94mm \\
    (i) & 5 & GT Heatmap & GT Proposal & - & - & - & - & 11.32mm \\
    \bottomrule
    \toprule
    & \# Views  & Backbone & CPN Size & AP$_{25}^{\text{rel}}$ & AP$_{50}^{\text{rel}}$ & AP$_{100}^{\text{rel}}$  & AP$_{150}^{\text{rel}}$ & MPJPE$^{\text{rel}}$\\
    \midrule
    (j) & 1 & ResNet-50 & $80\times 80 \times 20$  & 1.520 & 39.86 & 92.37 &  96.98 & 51.14mm\\
    \bottomrule
    \end{tabular}
    \caption{Ablation study on the Panoptic dataset. ``*'' means that CPN and PRN are trained on synthetic heatmaps. ``+'' means that CPN and PRN are trained and tested with different cameras. ``rel'' represents that we align the root joints of the estimated poses to the ground-truth.}
    \label{tab:ablative_panoptic}
\end{table*}

\paragraph{\textbf{Impact of Proposals}}
By comparing (b) and (h), we can see that replacing CPN by ground-truth proposals does not notably improve the results. The results suggest that the estimated proposals are already very accurate and more attention should be spent on improving the heatmaps and PRN. We do not compute APs when using ground-truth proposals because the confidence scores of all proposals are all set to be one.

\begin{figure*}
	\centering
	\includegraphics[width=4.7in]{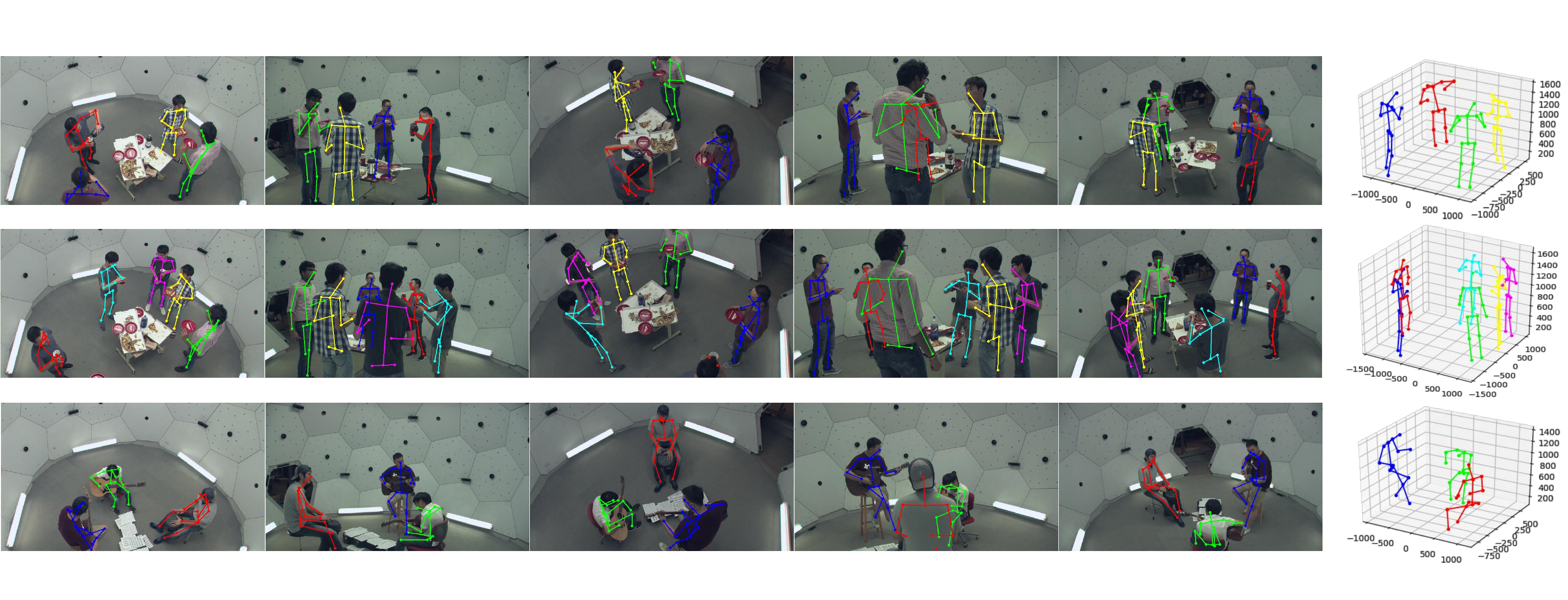}
	\caption{Estimated $3$D poses and their projections in images. The last column shows the estimated $3$D poses.
	}
	\label{fig:sample}
\end{figure*}

\paragraph{\textbf{Qualitative Study}}
We show the estimated $3$D poses of three examples in Figure \ref{fig:sample}. We can see that there are severe occlusions in the images of all camera views. However, by fusing the noisy and incomplete heatmaps from multiple cameras, our approach obtains more comprehensive features which allows us to successfully estimate the $3$D poses without bells and whistles. It is noteworthy that we do not need to associate $2$D poses in different views based on noisy observations by combining a number of sophisticated techniques. This significantly improves the robustness of the approach. Please see the supplementary video for more examples \footnote{https://youtu.be/qZAyHUzdpgw}.

Figure \ref{fig:badcase} (B) shows two examples where our approach did not obtain accurate estimations in the three-camera setup. In the first example, most joints of the lady can be seen from two of the three cameras and our approach accurately estimates the $3$D pose. However, the little child is only visible in the first view and, even in that view, many joints are actually occluded by its body. So the resulting $3$D pose has large errors. The second example is also interesting. The person is only visible in one view but, fortunately, most joints are visible. We can see that our approach estimates a $3$D pose which seems like a translated version of the ground-truth pose plotted in dashed lines. This is reasonable because there is ambiguity for $3$D pose estimation from a single image.

\paragraph{\textbf{Computational Complexity}}
It takes about $300$ms on a single Titan X GPU to estimate $3$D poses in a five-camera setup. In particular, $93$ms is spent on estimating heatmaps and $24$ms is spent on generating proposals. The time spent on regressing poses depends on the number of proposals (people). In particular, it takes about $46$ms to process one proposal. The inference time has the potential to be further reduced by using sparse $3$D convolutions \cite{yan2018second}.

\subsection{Comparison to the State-of-the-arts}
Table \ref{tab:campus_and_shelf} shows the results of the state-of-the-art methods on the Campus and the Shelf datasets in the top and bottom sections, respectively. On the Campus dataset, we can see that our approach improves PCP3D from $96.3\%$ of \cite{dong2019fast} to $96.7\%$ which is a decent improvement considering the already very high accuracy. As discussed in Section \ref{sec:datasets}, the PCP3D metric does not penalize false positive estimates. However, it is also meaningless to report AP scores because the GT pose annotations in this dataset are incomplete. So we propose to visualize and publish all of our estimated poses\footnote{https://youtu.be/AgDQFIlL5IM}. We find that our approach usually gets accurate estimates as long as joints are visible in at least two views. 

\begin{figure*}
	\centering
	\includegraphics[width=4.7in]{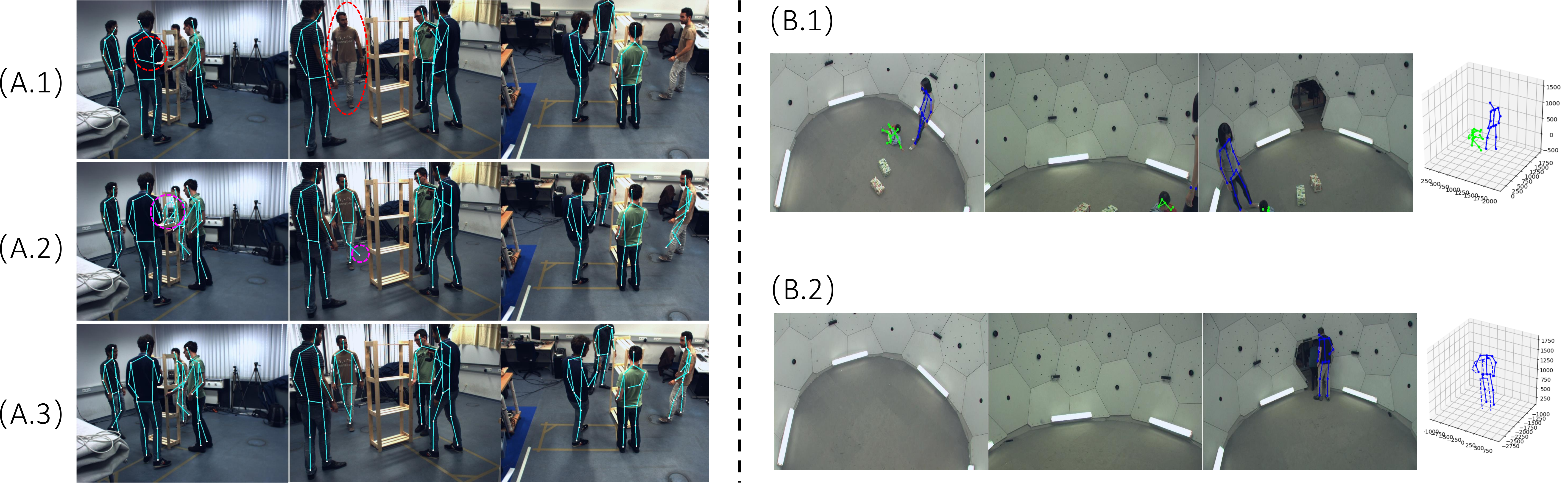}
	\caption{(A) shows the $3$D poses of ground-truth (A.1), estimated by \cite{dong2019fast} (A.2) and ours (A.3), respectively. The joints in the dashed circles represent the locations are incorrect. (B) shows two typical cases where our approach makes mistakes. The pose plotted by dashed lines in B.2 is the ground-truth.
	}
	\label{fig:badcase}
\end{figure*}

Our approach also achieves better results than \cite{dong2019fast} on the Shelf dataset. In particular, it gets fewer false positives. For example, in Figure \ref{fig:badcase} (A.2), there is a false positive pose in the pink dashed circle estimated by \cite{dong2019fast}. In contrast, our approach can suppress most false positives. We find that most errors of our approach are caused by inaccurate GT annotations. For example, as shown in the first column of Figure \ref{fig:badcase} (A.1), the GT joint locations within the red circle are incorrect. In summary, 66 out of the 301 frames have completely correct annotations and our approach gets accurate estimates on them.

\begin{table}[]
    \setlength{\tabcolsep}{10pt}
    \centering
    \begin{tabular}{r|cccc}
        \toprule
         Campus &  Actor 1 & Actor 2 & Actor 3 & Average \\
         \hline
         Belagiannis \etal \cite{belagiannis20143d} & 82.0 & 72.4 & 73.7 & 75.8\\
         Belagiannis \etal \cite{belagiannis2014multiple} & 83.0 & 73.0 & 78.0 & 78.0\\
         Belagiannis \etal \cite{belagiannis20153d} & 93.5 & 75.7 & 84.4 & 84.5\\
         Ershadi-Nasab \etal \cite{ershadi2018multiple} & 94.2 & 92.9 & 84.6 & 90.6\\
         Dong \etal \cite{dong2019fast} & 97.6 & 93.3 & 98.0 & 96.3\\
         Ours & 97.6 & 93.8 & 98.8 & \textbf{96.7} \\
         \hline
         \hline
          Shelf &  Actor 1 & Actor 2 & Actor 3 & Average \\
          \hline
          Belagiannis \etal \cite{belagiannis20143d} & 66.1 & 65.0 & 83.2 & 71.4\\
          Belagiannis \etal \cite{belagiannis2014multiple} & 75.0 & 67.0 & 86.0 & 76.0\\
          Belagiannis \etal \cite{belagiannis20153d} & 75.3 & 69.7 & 87.6 & 77.5\\
          Ershadi-Nasab \etal \cite{ershadi2018multiple} & 93.3 & 75.9 & 94.8 & 88.0\\
         Dong \etal \cite{dong2019fast} & 98.8 & 94.1 & 97.8 & 96.9\\
         Ours & 99.3 & 94.1 & 97.6 & \textbf{97.0}\\
         \bottomrule
    \end{tabular}
    \caption{Comparison to the state-of-the-art methods on the Campus and the Shelf datasets. The metric is PCP3D. }
    \label{tab:campus_and_shelf}
\end{table}

The previous works \cite{belagiannis20143d,belagiannis20153d,belagiannis2014multiple,dong2019fast} did not report numerical results on the large scale Panoptic dataset. We encourage future works to do so as in Table \ref{tab:ablative_panoptic} (b). We also evaluate our approach on the single person dataset Human3.6M \cite{ionescu2014human3}. The MPJPE of our approach is about $19$mm which is comparable to \cite{iskakov2019learnable}. We also visualize and publish our estimated poses \footnote{https://youtu.be/S6G3TXaBukw}.

\section{Conclusion}
we present a novel approach for multi-person $3$D pose estimation. Different from the previous methods, it only makes hard decisions in the $3$D space which allows to avoid the challenging association problems in the $2$D space. In particular, noisy and incomplete information of all camera views are warped to a common $3$D space to form a comprehensive feature volume which is used for $3$D estimation. The experimental results on the benchmark datasets validate that the approach is robust to occlusion which has practical values. In addition, the approach has strong generalization capability to different camera setups.

\clearpage

\bibliographystyle{splncs}
\bibliography{egbib}
\end{document}